\documentclass[conference]{IEEEtran}
\IEEEoverridecommandlockouts
\usepackage{cite}
\usepackage{amsmath,amssymb,amsfonts}
\usepackage{algorithm}
\usepackage{algorithmic}
\usepackage{graphicx}
\usepackage{textcomp}
\usepackage{xcolor}
\usepackage{url}
\def\BibTeX{{\rm B\kern-.05em{\sc i\kern-.025em b}\kern-.08em
    T\kern-.1667em\lower.7ex\hbox{E}\kern-.125emX}}
\begin{document}

\makeatletter
\newcommand\fs@norules{\def\@fs@cfont{\bfseries}\let\@fs@capt\floatc@ruled
  \def\@fs@pre{}%
  \def\@fs@post{}%
  \def\@fs@mid{\kern3pt}%
  \let\@fs@iftopcapt\iftrue}
\makeatother
\floatstyle{norules}
\restylefloat{algorithm}

\title{Vision Encoder-Decoder Models for AI Coaching}

\author{\IEEEauthorblockN{Dr. Jyothi S Nayak}
\IEEEauthorblockA{\textit{Department of Computer Science and Engineering} \\
\textit{B. M. S. College of Engineering}\\
Bangalore, India \\
jyothinayak.cse@bmsce.ac.in}
\and
\IEEEauthorblockN{Afifah Khan Mohammed Ajmal Khan}
\IEEEauthorblockA{\textit{Department of Computer Science and Engineering} \\
\textit{B. M. S. College of Engineering}\\
Bangalore, India \\
afifah.cs20@bmsce.ac.in}
\and
\IEEEauthorblockN{Chirag Manjeshwar}
\IEEEauthorblockA{\textit{Department of Computer Science and Engineering} \\
\textit{B. M. S. College of Engineering}\\
Bangalore, India \\
chirag.cs20@bmsce.ac.in}
\and
\IEEEauthorblockN{Imadh Ajaz Banday}
\IEEEauthorblockA{\textit{Department of Computer Science and Engineering} \\
\textit{B. M. S. College of Engineering}\\
Bangalore, India \\
imadh.cs20@bmsce.ac.in}
}
\maketitle

\begin{abstract}

This research paper introduces an innovative AI coaching approach by integrating vision-encoder-decoder models. The feasibility of this method is demonstrated using a Vision Transformer as the encoder and GPT-2 as the decoder, achieving a seamless integration of visual input and textual interaction. Departing from conventional practices of employing distinct models for image recognition and text-based coaching, our integrated architecture directly processes input images, enabling natural question-and-answer dialogues with the AI coach. This unique strategy simplifies model architecture while enhancing the overall user experience in human-AI interactions. We showcase sample results to demonstrate the capability of the model. The results underscore the methodology's potential as a promising paradigm for creating efficient AI coach models in various domains involving visual inputs. Importantly, this potential holds true regardless of the particular visual encoder or text decoder chosen. Additionally, we conducted experiments with different sizes of GPT-2 to assess the impact on AI coach performance, providing valuable insights into the scalability and versatility of our proposed methodology.
\end{abstract}


\section{Introduction}
Artificial Intelligence (AI) has achieved remarkable milestones, notably exemplified by the capabilities demonstrated by GPT-3 \cite{gpt3} in sustaining natural conversations, akin to human interactions, and acquiring extensive knowledge. This progress has opened avenues for practical applications, particularly in the domain of AI coach systems. These systems excel in delivering personalized experiences while ensuring privacy, leveraging AI's proficiency in learning and engaging in natural dialogue.

Acknowledging the vital role of visual context in coaching scenarios, especially when individuals seek feedback on activities like refining athletic form, we delve into the integration of vision encoder-decoder models inspired by the Transformer architecture \cite{transformer} in the domain of AI coaching. Departing from conventional methodologies, which typically entail a two-step process—firstly translating visual information into text and then employing a text-based model for coaching—we propose a more direct and efficient approach: a single unified model.

Our demonstration employs a vision transformer \cite{visionTransformer} as the encoder and GPT-2 \cite{gpt2} as the decoder, although acknowledging the versatility to utilize alternative visual encoders and text decoders. To illustrate the capabilities of our model, we present a demonstrative coaching task where users can input an image of a tic-tac-toe board, and pose questions referencing the image, e.g., "What is the best move to play?" While the simplicity of tic-tac-toe serves as a demonstrative task, the underlying methodology holds promise for more intricate applications.

In our experiments, our model exhibits proficiency in single-question/answering scenarios. Acknowledging the limitations imposed by our current hardware, we recognize the potential for extending the model's capabilities to engage in more extensive and nuanced conversations with the integration of more powerful text decoders.

Our methodology strategically capitalizes on the formidable pre-training capabilities embedded within both the visual encoder and text decoder. This approach not only alleviates the training workload but also amplifies the model's capacity to tackle a spectrum of coaching tasks with adeptness, as suggested in \cite{pretraining}.

While our current model serves as a demonstration, we envision future applications in sophisticated domains like AI-guided chess coaching. As computational resources advance, the proposed methodology holds promise for tackling more intricate coaching tasks, further enhancing the capabilities of AI coaching systems. This paper provides a foundational exploration of the integration of vision and text models, paving the way for future developments in the realm of AI coaching and beyond.

\section{Related Work}

Artificial intelligence (AI) coaching is a burgeoning field, attracting increasing attention in recent years. Several noteworthy research papers contribute to our understanding of this domain:

* Graßmann et al. explore the feasibility of AI coaching in their paper "Coaching With Artificial Intelligence: Concepts and Capabilities" \cite{grabmann}. While noting challenges in problem identification and individual feedback, they highlight AI coaching's potential in guiding clients and building alliances. The study provides insights for HRD professionals and suggests that, despite limitations, AI coaching's cost-effectiveness and broad reach could transform the coaching profession as a valuable HRD tool.

* Toshniwal et al. explore the domain of AI coaching in their paper titled "AI Coach for Badminton" \cite{toshniwal}. This research addresses the competitive world of sports, and show AI can help coach atheletes.

* Luo et al., in the paper "Artificial Intelligence Coaches for Sales Agents: Caveats and Solutions," \cite{toshniwal}delve into the application of AI coaches for training sales agents. Firms are increasingly leveraging AI coaches to enhance the skills of their sales teams. The authors present findings from a series of randomized field experiments, revealing intriguing insights into the nuances of this approach.

* Zhou et al., in "AI Coach for Battle Royale Games" \cite{zhou}, tackle the steep learning curve in the Battle Royale genre by introducing an AI Coach for Apex Legends players. Similar to our coaching concept, their algorithm uses player data to offer personalized suggestions for skill improvement. The study shows a 4-7

Research involving multi-modal input models, especially those incorporating text and image inputs, has influenced our model strategy. Research papers that have contributed to our understanding of this type of model include:

* Wang et al., in their paper "GIT: A Generative Image-to-text Transformer for Vision and Language" \cite{git}, introduce an approach to unify vision-language tasks, such as image/video captioning and question answering. The GIT model simplifies the architecture with a single image encoder and one text decoder, achieving state-of-the-art performance on 12 challenging benchmarks. This work significantly contributes to the understanding and advancement of generative models in the context of vision and language tasks, providing valuable insights for our own model development.

* Kim et al.'s "ViLT: Vision-and-Language Transformer Without Convolution or Region Supervision" \cite{vilt}, challenges traditional VLP methods by introducing a minimalistic model without convolution or region supervision. ViLT achieves remarkable speed gains (tens of times faster) while maintaining competitive downstream task performance. This work highlights the efficiency and expressive power benefits of a simplified visual input processing approach in VLP models.

* Wang et al.'s "Distilled Dual-Encoder Model for Vision-Language Understanding" \cite{wang} diverges from our model architecture by utilizing a dual-encoder instead of our visual encoder. Their DiDE framework aims to strike a balance between efficiency and performance, distilling knowledge from a fusion-encoder teacher model into a more efficient dual-encoder student model.

\section{Model Architecture} \label{modelSection}
\begin{figure*}[htbp]
\centerline{\includegraphics[width=16cm]{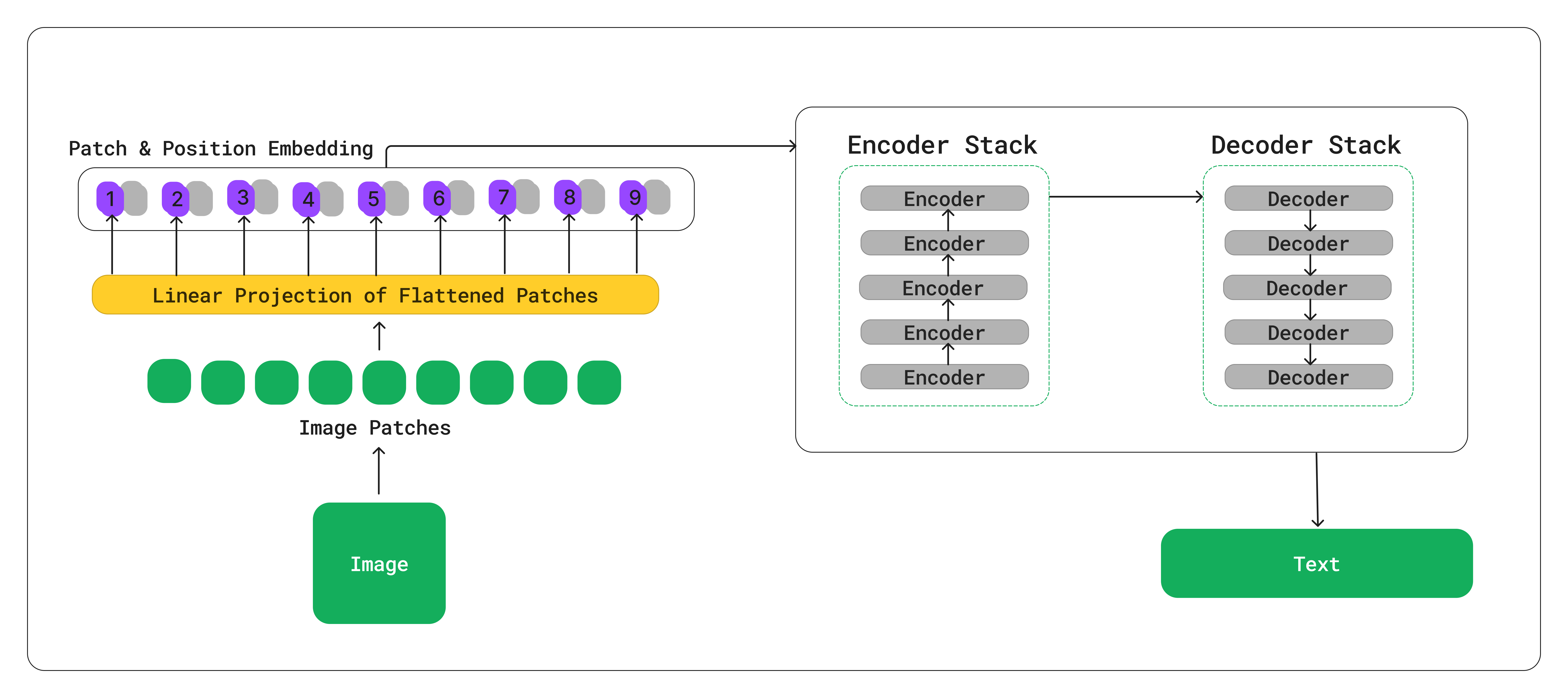}}
\caption{Model Architecture}
\label{fig1}
\end{figure*}
Figure \ref{fig1} shows the overall architecture of our model. Our model architecture is based on a encoder-decoder framework inspired by the Transformer architecture \cite{transformer}. The encoder is a pre-trained Vision Transformer (ViT) model \cite{visionTransformer}, which extracts high-level visual features from the input image. The decoder is a pre-trained GPT-2 model \cite{gpt2}, which generates text conditioned on the encoder's hidden states.

The ViT encoder is a transformer-based model that divides the input image into patches and extracts features from each patch using a self-attention mechanism. The self-attention mechanism allows the model to learn long-range dependencies in the image. The output of the ViT encoder is a sequence of hidden states, which represent the high-level visual features of the image.

The GPT-2 decoder is a transformer-based model that generates text one token at a time. The decoder is conditioned on the encoder's hidden states, which allows it to generate text that is relevant to the image.

The training process involves providing the input image to the encoder, transmitting the encoder's hidden states to the decoder, and training the decoder through the teacher-forcing algorithm. In this algorithm, during the training phase, the decoder is supplied with the correct text one token at a time, and the model learns to predict the subsequent token.

For the application of our model, we input an image of a tic-tac-toe board. To generate a meaningful response, we force the decoder to produce tokens representing the user's questions about the board using decoder-forcing. The consequently generated text from the decoder serves as the model's answer, taking context from both the visual features extracted from the tic-tac-toe board as well as the user's question. 

We harness the strength of pretraining by employing pre-trained versions for both the encoder and the decoder. This allows us to leverage the knowledge and features acquired from extensive datasets and tasks, providing a robust foundation before fine-tuning on our specific application.

\section{Training Dataset Generation} 
\label{datasetSection}

To train our model, we created a comprehensive dataset comprising all possible Tic-Tac-Toe boards, including both legal and illegal configurations, along with corresponding conversations related to the board. This dataset serves as the foundation for training the model to perform three key tasks: determining board validity, identifying the winner, and predicting the best move.

\subsection{Board Generation}

The first step involved generating all possible Tic-Tac-Toe boards using the \texttt{GenerateAll} algorithm (refer section \ref{algorithms}). This algorithm systematically generates configurations, including both legal and illegal configurations of the board.

\subsection{Image Representation}

For each generated Tic-Tac-Toe board, we utilized the Python Imaging Library (PIL) to create an image representation. The image encapsulates the visual layout of the board, enabling the model to interpret and learn from the game state visually.

\subsection{Task Outputs}

To facilitate supervised learning, we computed the expected outputs for three tasks:

\subsubsection{Board Validity}

The task of determining the validity of the board is handled by the \texttt{boardValidity} algorithm (refer section \ref{algorithms}. It assesses whether a given board is a valid Tic-Tac-Toe configuration.

\subsubsection{Winner Prediction}

The \texttt{getWinner} algorithm (refer section \ref{algorithms}) predicts the winner of a given board. It identifies whether 'X,' 'O,' or neither has won the game.

\subsubsection{Best Move Prediction}

The \texttt{bestMove} algorithm (refer section \ref{algorithms}) predicts the optimal move for a player in a specific board state.

\subsection{Conversation Generation}

The outputs of the aforementioned tasks are fed into the \texttt{ConversationGenerator} (refer section algorithm \ref{algorithms}). This algorithm processes the task outputs and generates text conversations that serve as training data for the model.

\subsection{Dataset Composition}

Our final dataset consists of pairs, each comprising a PIL-generated image of a Tic-Tac-Toe board and a corresponding text conversation for one of the mentioned tasks, generated by the \texttt{ConversationGenerator} algorithm. All possible combinations of the Tic-Tac-Toe board image and corresponding task conversations are present in the dataset. This pairing forms the input-output pairs for training the model. The image is fed into the model during training, and the conversation text is the expected output.

This comprehensive dataset ensures that the model is trained on diverse Tic-Tac-Toe scenarios, encompassing various board states and corresponding textual conversations.

\subsection{Algorithm Descriptions}\label{algorithms}

In this section, we provide detailed descriptions of the algorithms used in the dataset generation process.

\subsubsection{GenerateAll Algorithm}

\begin{algorithm}[H]
 \caption{GenerateAll Algorithm}
 \begin{algorithmic}[1]
 \renewcommand{\algorithmicrequire}{\textbf{Input:} None}
 \renewcommand{\algorithmicensure}{\textbf{Output:} }
 \ENSURE  List of all possible Tic-Tac-Toe boards
  \STATE \textbf{Function} generateAll():
  \STATE \quad \textbf{Initialization} :
  \STATE \quad \quad Create a 3x3 board filled with empty spaces
  \STATE \quad \quad Initialize an empty list \textit{result}
  \STATE \quad \textbf{Backtrack Function} :
  \STATE \quad \quad \textbf{Function} backtrack(board, row, col, result):
  \STATE \quad \quad \quad \textbf{Base Case} :
  \STATE \quad \quad \quad \quad If the current row is 3, append the board to \textit{result} and return
  \STATE \quad \quad \quad \textbf{Calculate Next Row and Column} :
  \STATE \quad \quad \quad \quad Calculate \textit{next\_row} and \textit{next\_col} based on the current row and column
  \STATE \quad \quad \textbf{Try All Symbols} :
  \STATE \quad \quad \quad For each symbol in ['\_', 'X', 'O']:
  \STATE \quad \quad \quad \quad Set the current cell on the board to the symbol
  \STATE \quad \quad \quad \quad Recursively call \textit{backtrack} with the updated board and positions
  \STATE \quad \textbf{Main Algorithm} :
  \STATE \quad \quad Create a 3x3 board filled with empty spaces
  \STATE \quad \quad Initialize an empty list \textit{result}
  \STATE \quad \quad Call \textit{backtrack} with the initial board, start at (0, 0)
  \STATE \quad \quad \textbf{Return} the \textit{result} as the list of all possible Tic-Tac-Toe boards
  \RETURN $result$
 \end{algorithmic}
\end{algorithm}

This algorithm employs a recursive approach to systematically explore all possible Tic-Tac-Toe board configurations. The recursive function generates valid and invalid states, ensuring a comprehensive dataset that covers a wide range of game scenarios.

\subsubsection{Image Generation Algorithm}

\begin{algorithm}[H]
\caption{Image Generation Algorithm}
\begin{algorithmic}[1]
 \renewcommand{\algorithmicrequire}{\textbf{Input:}}
 \renewcommand{\algorithmicensure}{\textbf{Output:}}
\REQUIRE Tic-Tac-Toe board configuration
\ENSURE PIL-generated image of the Tic-Tac-Toe board
\STATE \textit{Initialization}:
\STATE Initialize an empty PIL image with a white background
\STATE \textit{Draw Symbols}:
\STATE Use PIL drawing functions to represent 'X' and 'O' on the image
\STATE Draw grid lines to separate the board
\RETURN PIL-generated image
\end{algorithmic}
\end{algorithm}

This algorithm converts a given Tic-Tac-Toe board configuration into a visual representation using the Python Imaging Library (PIL).

\subsubsection{boardValidity Algorithm}

\begin{algorithm}[H]
 \caption{boardValidity Algorithm}
 \begin{algorithmic}[1]
 \renewcommand{\algorithmicrequire}{\textbf{Input:}}
 \renewcommand{\algorithmicensure}{\textbf{Output:}}
 \REQUIRE Tic-Tac-Toe board configuration
 \ENSURE True if the board is a valid Tic-Tac-Toe configuration, False otherwise
 \STATE Initialize counters for 'X' and 'O' to 0
 \FOR{each row in the board}
    \FOR{each cell in the row}
        \IF{cell is 'X'}
            \STATE Increment 'X' counter
        \ELSIF{cell is 'O'}
            \STATE Increment 'O' counter
        \ENDIF
    \ENDFOR
 \ENDFOR
 \IF{'O' count is greater than 'X' count or 'O' count is less than 'X' count minus 1}
    \RETURN False
 \ENDIF
 \STATE Define win conditions for Tic-Tac-Toe
 \STATE Initialize a variable 'won' to False
 \FOR{each win condition}
    \STATE Extract symbols from cells in the win condition
    \IF{symbols form a winning combination for 'X' or 'O'}
        \IF{already 'won' is True}
            \RETURN False
        \ENDIF
        \STATE Set 'won' to True
    \ENDIF
 \ENDFOR
 \RETURN True
 \end{algorithmic}
\end{algorithm}

This algorithm takes a Tic-Tac-Toe board configuration as input and checks its validity. It does so by counting the number of 'X' and 'O' symbols, ensuring that 'X' starts first and that there are no multiple wins. It checks the rows, columns, and diagonals for winning combinations and returns True if the board is valid, indicating a proper Tic-Tac-Toe game state. Otherwise, it returns False.

\subsubsection{getWinner Algorithm}

\begin{algorithm}[H]
 \caption{getWinner Algorithm}
 \begin{algorithmic}[1]
 \renewcommand{\algorithmicrequire}{\textbf{Input:}}
 \renewcommand{\algorithmicensure}{\textbf{Output:}}
 \REQUIRE Tic-Tac-Toe board configuration
 \ENSURE The winner: 'X' or 'O', or '\textbackslash' if there's no winner yet
 \STATE Call \textbf{boardValidity} to check if the board is valid
 \IF{board is not valid}
    \RETURN False
 \ENDIF
 \STATE Define win conditions for Tic-Tac-Toe
 \STATE Initialize a variable 'won' to False
 \FOR{each win condition}
    \STATE Extract symbols from cells in the win condition
    \IF{symbols form a winning combination for 'X' or 'O'}
        \RETURN The winner: 'X' or 'O'
    \ENDIF
 \ENDFOR
 \RETURN '\textbackslash'
 \end{algorithmic}
\end{algorithm}

This algorithm determines the winner in a Tic-Tac-Toe game by first validating the board and then analyzing rows, columns, and diagonals for winning scenarios. It returns the winner or '\_' when there is no winner.

\subsubsection{bestMove Algorithm}

\begin{algorithm}[H]
 \caption{bestMove Algorithm}
 \begin{algorithmic}[1]
 \renewcommand{\algorithmicrequire}{\textbf{Input:}}
 \renewcommand{\algorithmicensure}{\textbf{Output:}}
 \REQUIRE Tic-Tac-Toe board configuration, current player ('X' or 'O')
 \ENSURE The best move for the current player
 \IF{current player is 'X'}
    \STATE Initialize bestScore to negative infinity
 \ELSE
    \STATE Initialize bestScore to positive infinity
 \ENDIF
 \STATE Initialize bestMove to None
 \FOR{each empty cell in the board}
    \STATE Place the current player's symbol in the empty cell
    \STATE Calculate the score for the current move using the minimax algorithm
    \IF{current player is 'X' and score is greater than bestScore}
        \STATE Update bestScore to the score
        \STATE Update bestMove to the current move
    \ELSIF{current player is 'O' and score is less than bestScore}
        \STATE Update bestScore to the score
        \STATE Update bestMove to the current move
    \ENDIF
    \STATE Remove the current player's symbol from the cell
 \ENDFOR
 \RETURN bestMove
 \end{algorithmic}
\end{algorithm}

This algorithm employs the minimax algorithm to predict the optimal move for a player in a given Tic-Tac-Toe board state. It considers all possible future moves to make strategic decisions.

\subsubsection{ConversationGenerator Algorithm}

\begin{algorithm}[H]
\caption{ConversationGenerator Algorithm}
\begin{algorithmic}
\renewcommand{\algorithmicrequire}{\textbf{Input:}}
\renewcommand{\algorithmicensure}{\textbf{Output:}}
\REQUIRE Task type, Tic-Tac-Toe board configuration
\ENSURE Question and Answer strings
\STATE Initialize question to an empty string
\STATE Initialize answer to an empty string
\IF{task\_type is 'next\_move'}
  \STATE Generate a question about the next move for the current player
  \STATE Calculate the best move using the 'bestMove' algorithm provided earlier
  \STATE Generate an answer with the best move
\ELSIF{task\_type is 'winner'}
  \STATE Generate a question about the winner of the game
  \STATE Check the board configuration for a winner ('X' or 'O') or a draw
  \STATE Generate an answer with the winner or a draw
\ELSIF{task\_type is 'valid\_move'}
  \STATE Generate a question about the validity of a specific move
  \STATE Check if the provided move is valid on the board
  \STATE Generate an answer indicating whether the move is valid or not
\ENDIF
\RETURN question, answer
\end{algorithmic}
\end{algorithm}

This algorithm synthesizes the results from the board validity, winner prediction, and best move prediction tasks into a coherent text conversation. The question and answer strings are not hard-coded to encourage generalization and avoid hard-coding language.

\section{Experimental Setup And Evaluation}

\begin{figure*}[htbp]
\centerline{\includegraphics[width=20cm]{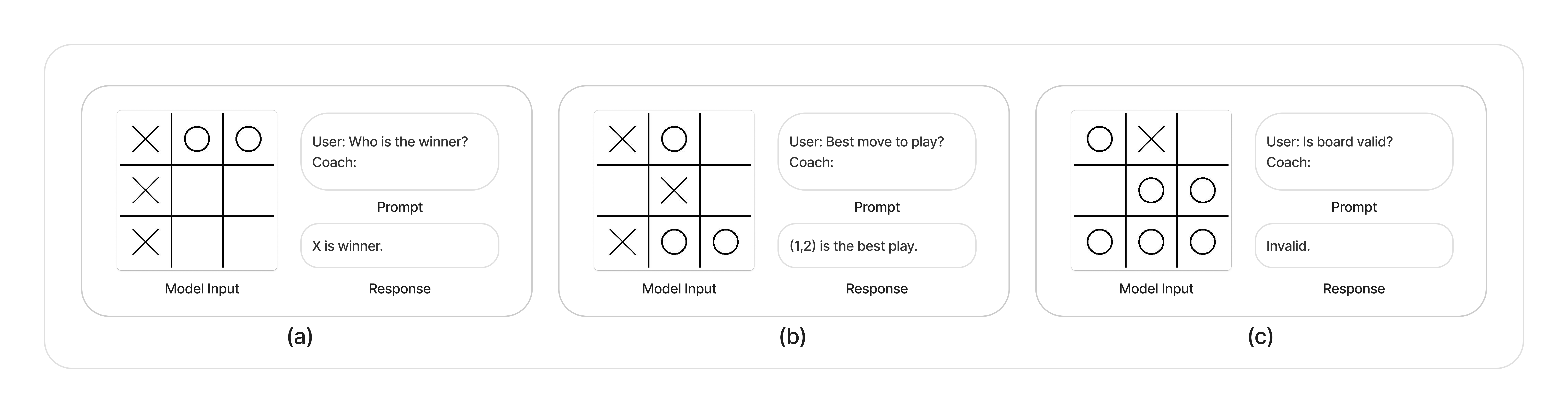}}
\caption{Sample outputs for tasks (a) Board Validity, (b) Winner Prediction, and (c) Best Move Prediction.}
\label{fig2}
\end{figure*}

Our experimental setup and training process are detailed below:

\subsection{Dataset}

We utilized a generated dataset (refer section \ref{datasetSection}) comprising 59,049 samples. To ensure robust model training, we divided the dataset into 99\% for training and 1\% for validation. Manual validation was conducted to assess model performance.

\subsection{Training Environment}

The training procedure was accelerated using a dedicated server optimized for high-performance tasks, equipped with an NVIDIA A100 40GB GPU.

\subsection{Training Configuration}

The model was trained for 3 epochs with varying sizes of GPT-2: 117M parameters, 345M parameters, and 774M parameters. We employed the Adam optimizer with a learning rate set to $1 \times 10^{-5}$. 

\subsection{Evaluation}

A glance at the results depicted in Figure \ref{fig2} showcases the promising potential of this methodology. During inference, the image is fed into the model, while the user's question is observed by decoder-forcing the prompt, as described in section \ref{modelSection}.

A notable observation becomes evident when analyzing the outcomes of the three iterations of models we trained with varying GPT-2 sizes. The evaluation results consistently reveal expected improvements in both accuracy and language fluency as the text decoder grows in size. This suggests that the dataset is sufficiently rich in providing comprehensive context. Furthermore, it implies that with a sufficient dataset, a larger text decoder equips the model with the capacity to derive more profound insights from the training data. This underscores the promising future prospects of this methodology.

\section{Conclusion}
In this paper, we presented a novel approach to AI coaching systems that seamlessly integrates vision and text models through a unified Transformer-based architecture. Our model, combining a Vision Transformer (ViT) as the encoder and GPT-2 as the decoder, demonstrated proficiency in addressing coaching tasks, as exemplified by the tic-tac-toe board scenario. The model's ability to provide meaningful responses to user queries based on visual information showcases the potential of our methodology.

The training process, leveraging pre-trained models for both the encoder and decoder, proved effective in harnessing the strength of pretraining, enhancing the model's capacity to handle a spectrum of coaching tasks. Our experiments highlighted the impact of text decoder size on performance, indicating that a larger decoder, given a sufficient dataset, can lead to improved accuracy and language fluency.

Our comprehensive dataset, encompassing diverse Tic-Tac-Toe scenarios, served as a robust foundation for model training. The generated dataset, consisting of 59,049 samples, facilitated the training of the model to perform key tasks such as determining board validity, predicting the winner, and suggesting the best move.

While our current model serves as a demonstration, we envision promising future applications in sophisticated domains, such as AI-guided chess coaching, as computational resources continue to advance. The proposed methodology opens avenues for tackling more intricate coaching tasks, thereby enhancing the capabilities of AI coaching systems.

In conclusion, our research provides a foundational exploration of the integration of vision-encoder decoder in the realm of AI coaching. The promising results and future prospects outlined in this paper pave the way for further developments and advancements in AI coaching systems and beyond.

\bibliographystyle{IEEEtran}
\bibliography{bibliography}

\end{document}